\documentclass[conference]{IEEEtran}
\IEEEoverridecommandlockouts
\usepackage{cite}
\usepackage{amsmath,amssymb,amsfonts}
\usepackage{algorithmic}
\usepackage{graphicx}
\usepackage{textcomp}
\usepackage{xcolor}
\usepackage{booktabs}
\usepackage{multirow}
\usepackage{enumitem}
\usepackage{multicol}
\usepackage{amssymb}
\usepackage{bbding}
\usepackage{wasysym}
\usepackage{colortbl}
\usepackage{algorithm}
\usepackage{setspace}
\usepackage{verbatim}
\usepackage{hyperref}

\def\BibTeX{{\rm B\kern-.05em{\sc i\kern-.025em b}\kern-.08em
    T\kern-.1667em\lower.7ex\hbox{E}\kern-.125emX}}
\begin{document}

\title{LLaVA-SG: Leveraging Scene Graphs as Visual Semantic Expression in Vision-Language Models}

\author{
    \IEEEauthorblockN{Jingyi Wang\textsuperscript{1}, Jianzhong Ju\textsuperscript{2}, Jian Luan\textsuperscript{2}, Zhidong Deng\textsuperscript{1*}\thanks{\textsuperscript{*}Corresponding author.}
    }
    \IEEEauthorblockA{\textsuperscript{1} Department of Computer Science and Technology, Tsinghua University, Beijing, China}
    \IEEEauthorblockA{\textsuperscript{2} Xiaomi AI Lab, Beijing, China}
    \IEEEauthorblockA{\{wang-jy20, michael\}@mails.tsinghua.edu.cn, \{jujianzhong, luanjian\}@xiaomi.com}
}

\maketitle

\begin{abstract}
Recent advances in large vision-language models (VLMs) typically employ vision encoders based on the Vision Transformer (ViT) architecture. The division of the images into patches by ViT results in a fragmented perception, thereby hindering the visual understanding capabilities of VLMs. 
In this paper, we propose an innovative enhancement to address this limitation by introducing a Scene Graph Expression (SGE) module in VLMs. This module extracts and structurally expresses the complex semantic information within images, thereby improving the foundational perception and understanding abilities of VLMs. 
Extensive experiments demonstrate that integrating our SGE module significantly enhances the VLM's performance in vision-language tasks, indicating its effectiveness in preserving intricate semantic details and facilitating better visual understanding. 
\end{abstract}


\section{Introduction}
\label{sec:intro}

\begin{figure}
  \centering
  \includegraphics[width=\columnwidth]{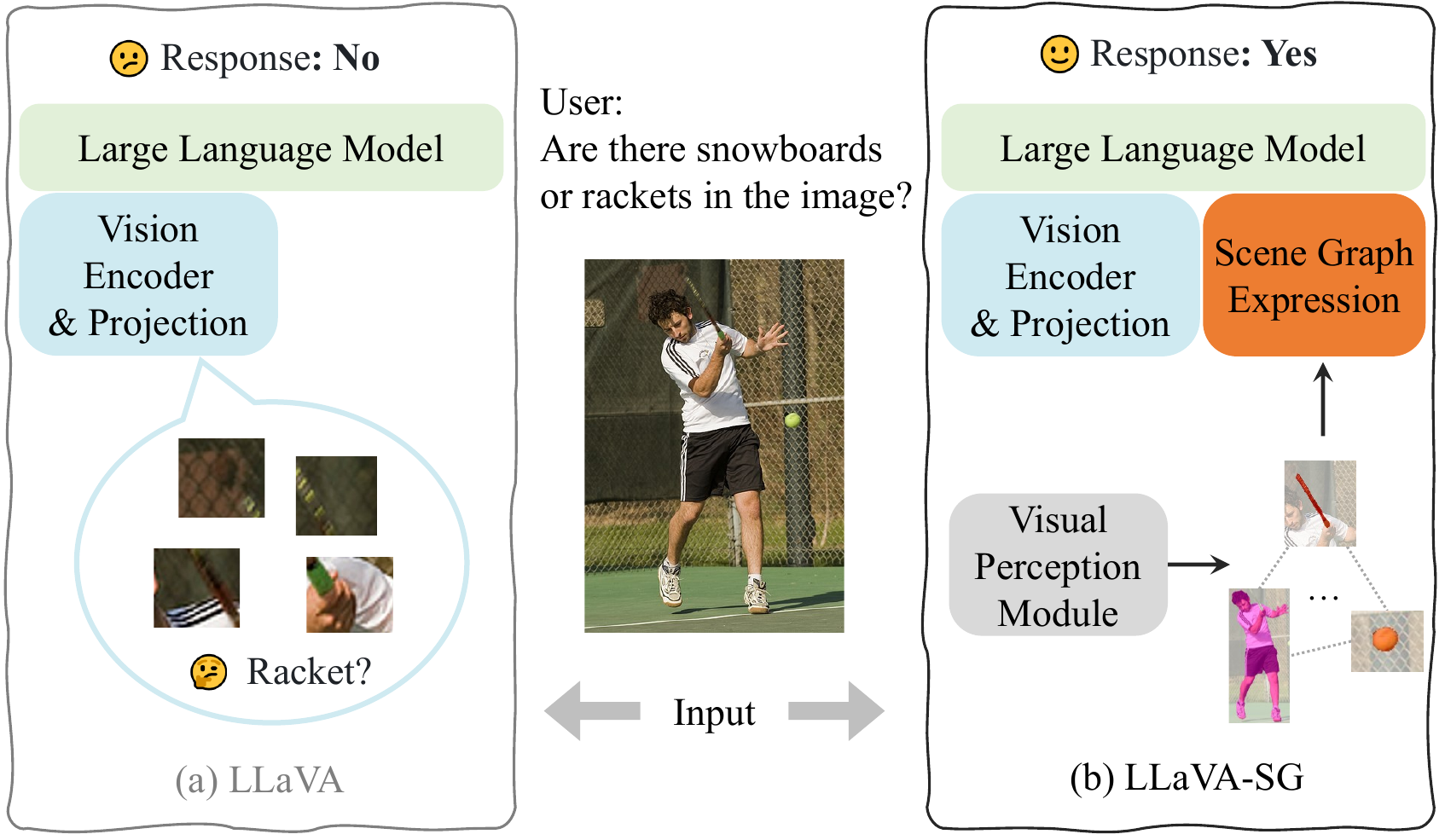}
  \vspace{-0.5cm}
  \caption{The illustration of the difference between (a) the baseline method Large Language and Vision Assistant (LLaVA)\cite{liu2024llava} and (b) our LLaVA-SG model. As a complement to the baseline method of dividing images into patches, our LLaVA-SG leverages scene graphs as the expression of visual semantic within images.}
  \vspace{-0.3cm}
  \label{fig:motivation}
\end{figure}

Large vision-language models (VLMs) integrate data from visual and language modalities, enabling comprehensive multimodal understanding \cite{herzig2023sgvl,yang2024icassp,zhang2024mllmsurvey}. 
With images and query text as input, VLMs can answer queries by incorporating the visual information.
However, most VLMs utilize the Vision Transformer (ViT) \cite{dosovitskiy2020vit} as their visual backbone, which results in perceiving images as fragmented patches, as shown in Fig.~\ref{fig:motivation}(a) with LLaVA as the VLM baseline\cite{liu2024llava}. This fragmented approach fails to preserve the intrinsic semantic information in images, thus limiting the VLMs' visual comprehension capabilities \cite{liu2024llava}. To address this limitation, we propose using scene graphs, which express the objects in a scene and the relationships between them to retain and structurally express the complex semantic information in images, as shown in Fig.~\ref{fig:motivation}(b). In this way, the visual perception and understanding capabilities of VLMs are enhanced with the proposed Scene Graph Expression (SGE) module.

\begin{figure*}[htbp]
\centering
\includegraphics[width=\textwidth]{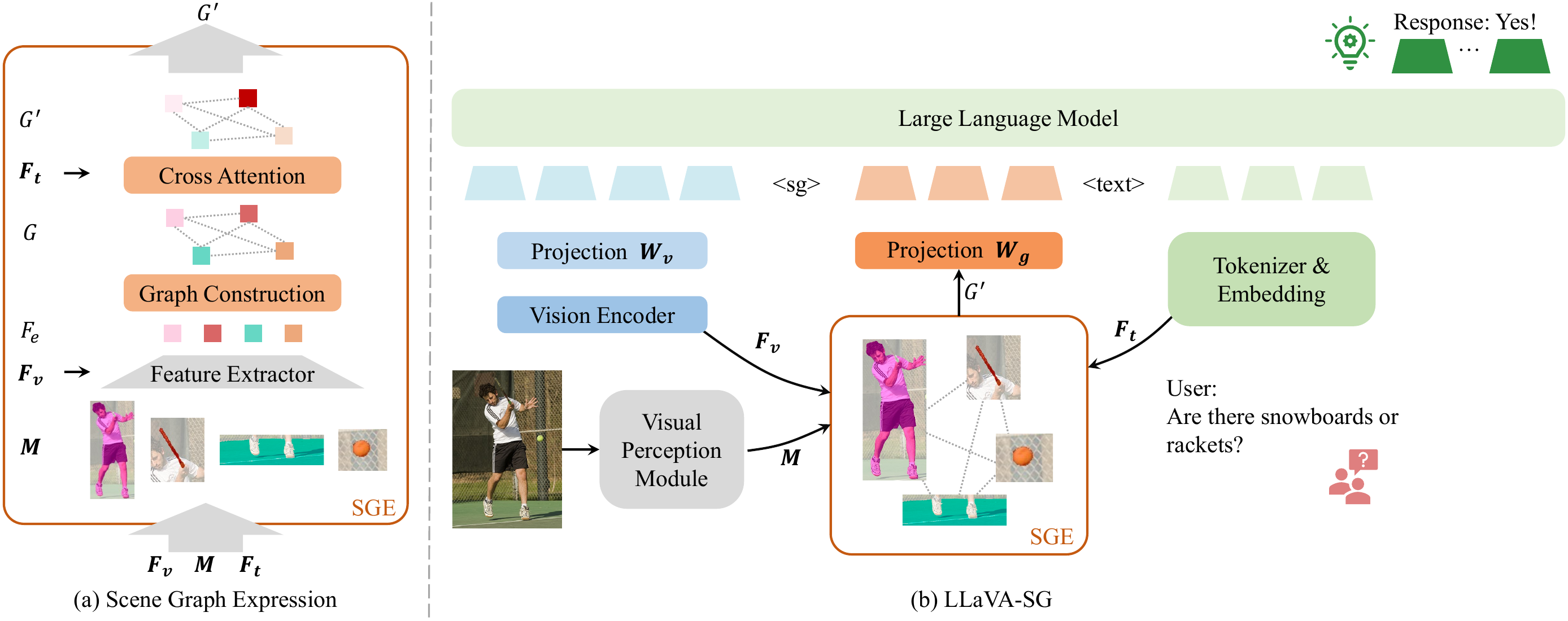}
\caption{The structure of the proposed Scene Graph Expression module and the LLaVA-SG framework.}
\vspace{-2pt}
\label{fig:model}
\end{figure*}

Improving the perceptual capabilities of VLMs has been the focus of several research efforts. 
For instance, Lyrics \cite{lu2023lyrics} introduces a visual refiner to incorporate semantic-aware visual objects. IVE \cite{he2024ive} integrates multitask encoders and visual tools into existing models. Several studies have also explored the use of scene graphs in VLMs \cite{herzig2023sgvl,mitra2024ccot}. 
However, training directly on scene graph data or fine-tuning on scene graph text has not yielded optimal results \cite{herzig2023sgvl,mitra2024ccot}.

To achieve the structured expression of visual semantic information without directly training on scene graph data, we designed the Scene Graph Expression module. Specifically, we use pretrained visual models to extract entities from images, preserving semantic information at the entity level rather than the patch level. Next, we construct a scene graph using these entities and perform message passing between the nodes in the scene graph. Building on LLaVA \cite{liu2024llava}, we construct LLaVA-SceneGraph (LLaVA-SG), a model that incorporate SGE module to enhance the foundational perception and understanding capabilities of VLMs. 
To enhance visual understanding, we train LLaVA-SG to classify visual relationships between the extracted entities, avoiding the catastrophic forgetting that might occur with direct training on scene graph data. Our contributions are threefold:
\begin{itemize}
    \item We design a Scene Graph Expression (SGE) module to extract and structurally express the intrinsic semantic information in images.
    \item We incorporate SGE into VLM, resulting in the LLaVA-SG model to enhance the foundational visual perception and understanding capabilities. 
    \item We conduct extensive evaluations of the trained LLaVA-SG model on multiple benchmarks, demonstrating that the integration of SGE significantly improves visual perception and understanding capabilities of VLM.
\end{itemize}

\section{Method}

In this section, we first introduce the scene graph construction to preserve and express complex semantic information within images.
Next, we detail the integration of SGE module into standard VLM models, forming the LLaVA-SG model.
Finally, we present the training strategies devised for LLaVA-SG. An overview of our approach is shown in Fig.~\ref{fig:model}(b).

\subsection{Semantic Information Expression}
\subsubsection{Visual Entity Extraction}
To achieve the structured expression of semantic information in a scene, we first extract the entities within the scene. Specifically, in the Visual Perception Module in Fig.~\ref{fig:model}(b), we first perform image tagging on the input to identify the categories of the entities it contains. Taking the input image in Fig.~\ref{fig:model}(b) as an example, the image tagging process will output tags such as "man", "racket", "tennis ball", and "sports field". The object detection module then detects the bounding boxes of these entities based on the tagging results. The bounding boxes are rectangular boxes that represents the position and size of each entity. Next, we conduct semantic segmentation within the bounding boxes to obtain the pixels comprising each entity, achieving pixel-level semantic understanding with the help of segmentation masks.
We denote the segmentation masks as $M \in \mathbb{R}^{N \times h_m \times w_m}$, where $N$ is the number of segmented entities. $h_m$ and $w_m$ denote the height and width of masks.

To express the semantic information in images, we design the Scene Graph Expression module, as depicted in Fig.~\ref{fig:model}(a). The segmentation masks $M$ serve as the input of the SGE module representing entities in the input image.

\begin{table*}[htbp] \centering
\caption{Comparison with multiple VLMs across multiple benchmarks including VQA-v2-test \cite{goyal2017vqav2}, GQA \cite{hudson2019gqa}, ScienceQA-IMG \cite{lu2022scienceqa}, MMBench \cite{liu2023mmbench}, MMBench-Chinese \cite{liu2023mmbench} and POPE \cite{li2023pope}.}  
\label{tab:result}
\resizebox{\textwidth}{!}{
\begin{tabular}{ll|cccccc|cccccc}
\toprule
\multirow{2}{*}{Model} & \multirow{2}{*}{LLM} & \multicolumn{6}{c|}{Benchmarks} & \multicolumn{6}{c}{Fine-Grained Ability from MMBench} \\  & & \makebox[23pt][c]{VQA\textsuperscript{v2}} & \makebox[23pt][c]{GQA} & \makebox[23pt][c]{SQA\textsuperscript{I}} & \makebox[23pt][c]{MMB} & \makebox[23pt][c]{MMB\textsuperscript{CN}} & \makebox[23pt][c]{POPE} & LR & AR & RR & FP-S & FP-C & CP \\ \midrule
BLIP-2 \cite{li2023blip2} & Vicuna-7B & 65.0 & 41 & 61 & - & - & 85.3 & - & - & - & - & - & - \\
mPlug-OWL2 \cite{ye2024mplug} & LLaMA2-7B & 79.4 & 56.1 & 68.7 & 64.5 & 36.2 & - & 32.2 & 72.4 & 60.9 & 68.6 & 60.1 & 79.4 \\
InstructBLIP \cite{instructblip} & Vicuna-7B & - & 49.2 & 60.5 & 36.0 & 23.7 & - & 21.6 & 47.4 & 22.5 & 33.0 & 24.4 & 41.1 \\
QwenVL \cite{bai2023qwen} & Qwen-7B & 78.8 & 59.3 & 67.1 & 38.2 & 7.8 & - & 9.8 & 43.1 & 30.3 & 32.9 & 27.9 & 36.4 \\
QwenVL-chat \cite{bai2023qwen} & Qwen-7B & 78.2 & 57.5 & 68.2 & 61.8 & 56.7 & - & 40.5 & 74.3 & 47.9 & 66.3 & 46.2 & 72.8 \\
LLaVA-1.5 \cite{liu2024llava1_5} & Vicuna-7B & 78.5 & 62.0 & 66.8 & 64.2 & 57.6 & 85.9 & 33.1 & 70.4 & 58.3 & 65.9 & 55.2 & 77.4 \\ \midrule
\rowcolor{gray!10} LLaVA-SG-7B & Vicuna-7B & 79.2 & 63.5 & 68.7 & 68.0 & 58.7 & 86.7 & 39.8 & 70.3 & \textbf{68.7} & 69.6 & 59.4 & 80.1 \\
\rowcolor{gray!10} LLaVA-SG-13B & Vicuna-13B & \textbf{80.1} & \textbf{63.6} & \textbf{71.1} & \textbf{68.7} & \textbf{61.7} & \textbf{86.9} & 41.5 & 70.4 & 60.0 & \textbf{73.4} & \textbf{61.5} & \textbf{80.7} \\ \bottomrule
\end{tabular}
}
\vspace{-0.2cm}
\end{table*}

\subsubsection{Scene Graph Expression}
\label{sec:sge}
We introduce the Scene Graph Expression (SGE) module to structurally preserve and express the semantic information in a scene at the entity level. A scene graph is a graphical representation that captures and expresses the entities and their relationships in a scene. 
Scene graphs aid in visual comprehension of images and videos, and facilitate scene understanding \cite{wang2023sil}. Entities in a scene are represented as entities in a scene graph.
Additionally, we introduce the prompt feature to activate key nodes in the scene graph. Next, we detail the steps involved in the SGE module.

To construct the scene graph for the input image, we first extract the visual features of the entities within the image to represent the nodes. Specifically, we use $F_v \in \mathbb{R}^{ d_e \times h_v \times w_v}$ to denote the visual features of the input image obtained from the pretrained vision encoder, where $d_e$ is the dimension of $F_v$. $h_v$ and $w_v$ denote the shape of the feature map. Following \cite{zou2024seem}, we extract the features of points in the masks $M$ from $F_v$ using bilinear interpolation. Then, for each mask in $M$, we average the features of the points it contains to obtain the feature representation of the mask. The features of masks are collectively denoted as $F_e \in \mathbb{R}^{ N \times d_e}$, representing the visual features of the corresponding entities. 
Second, we construct the scene graph $G$ with these $N$ entities as nodes of the graph and $F_e$ as the node features. Edges among the nodes depict their relationships. We perform message passing among the $N$ nodes to implicitly encode the relationship information between nodes.
Third, to make the scene graph expression adaptive, we utilize the prompt feature to activate the key nodes within $G$. We use $F_t \in \mathbb{R}^{ N_t \times d_t}$ to denote the prompt feature where $N_t$ is the number of prompt tokens, and $d_t$ is the dimension of the prompt feature.
Specifically, we adopt the attention mechanism to inject the prompt feature $F_t$ into $G$, highlighting the nodes in $G$ that are relevant to the prompt feature. The updated scene graph is denoted as $G^\prime$, which is the output of the SGE module, representing the structured semantic information contained in the input image.

\subsection{Scene Graph Expression in VLM}
To address the degradation of complex semantic information in VLMs caused by the processing of images as set of patches, we introduce the SGE module into the VLM, as shown in Fig.~\ref{fig:model}(b). 
Using LLaVA \cite{liu2024llava1_5} as the base framework, the image is input into the pretrained vision encoder, and the obtained visual features are used as the input of SGE module, that is, $F_v$. Utilizing the pretrained visual perception module, we obtain the semantic segmentation masks of the input image. 
The input text is tokenized, and the embeddings are used as the prompt feature for SGE module, denoted as $F_t$. Together, $F_v$, $M$, and $F_t$ serve as inputs to the SGE module, constructing the scene graph $G^\prime$ activated by the prompt feature. 
Similar to the projection layer $W_v$ after the pretrained vision encoder, we apply a trainable projection layer $W_g$ to convert the nodes in $G^\prime$ into language embedding tokens. With the same dimensionality, the converted scene graph tokens are then fed into the Large Language Model, along with the visual tokens and the text tokens. We add two special tokens to the LLM, namely \texttt{<sg>} and \texttt{<text>}, which are inserted before and after the scene graph tokens, respectively. Consequently, the sequence of these tokens is fed into the LLM to generate language response tokens as follows:
\begin{equation}
    \centering
    H_a = LLM ( [ W_v \cdot F_v, \texttt{<sg>}, W_g \cdot G^\prime, \texttt{<text>}, F_t ]),
\end{equation}
where $H_a$ denotes the output of the LLM, serving as the response to the input image and text.

\subsection{Training}
Following LLaVA \cite{liu2024llava}, we train our LLaVA-SG on the prediction tokens of the LLM using an auto-regressive training objective. Building on LLaVA's original two-stage training, we insert an additional training stage specifically for our SGE module. The training of LLaVA-SG thus includes the following three stages:

\textbf{Visual Feature Alignment.} We initialize the image encoder and the LLM of LLaVA-SG with pretrained weights and keep them frozen. We do not use the SGE module at this stage. Only the visual projection layer $W_v$ is trainable. 
The dataset of 558K LAION-CCSBU image-text pairs\cite{li2022blip,liu2024llava} is adopted for training in this stage. 

\textbf{SGE training.} In the second stage, we focus training the SGE module and the corresponding projection layer $W_g$. The visual encoder, visual projection layer, and LLM are all frozen during this stage. 
Considering that a scene graph contains both entities and the relationships between them, we endow the SGE module with the ability to preserve and express visual semantic information from both perspectives. For entity recognition, we use the pretrained visual perception module. For expressing the relationships between entities, we train the SGE module and the projection layer $W_g$ on visual relationship understanding datasets. 
Specifically, the visual relationship understanding datasets we use are derived from two sources. First, existing fine-grained visual understanding datasets, such as Visual Genome \cite{vg} and Open Image V6 \cite{oiv6}, are reformatted into the visual question-answering format. However, the relationships in these datasets are limited and not open-vocabulary. Therefore, we also construct an open-vocabulary visual relationship understanding dataset based on the large-scale visual grounding dataset GRIT \cite{peng2023grit}, utilizing GPT-4v to produce the data.

\textbf{Fine-tuning.} In the third stage, only the visual encoder remains frozen, while all other parameters are trained in LLaVA-SG. We use the 665K image-text instruction dataset from LLaVA-1.5 \cite{liu2024llava1_5}, which contains diverse instructions for fine-tuning in this stage.

\section{Experiments}

\subsection{Experimental Setup}
In LLaVA-SG, we adopt CLIP ViT-L/14@336p \cite{dosovitskiy2020vit, radford2021clip} as the vision encoder and Vicuna \cite{chiang2023vicuna} as the LLM. The pretrained visual perception module includes RAM \cite{zhang2024ram} for image tagging, Grounding-DINO \cite{groundingdino} for detection, and SAM \cite{kirillov2023sam} for semantic segmentation. The message passing and attention mechanism in the SGE module are implemented using lightweight transformers. In the first stage, the learning rate for the trainable parameters is set to $2e-3$. In the second and third stages, we use a learning rate of $2e-5$.

\begin{table}[tbp]\centering
\caption{Ablation study for the SGE module. COCO refers to the COCOcap-val-2017 task. Rel refers to the test sets of Visual Genome \cite{vg} and Open Images \cite{oiv6}.}  
\label{tab:ablationSGE}
\resizebox{\columnwidth}{!}{
\begin{tabular}{ccc|cccc}
\toprule
SG & MP & Prompt & \makebox[30pt][c]{GQA} & \makebox[30pt][c]{MMB} & \makebox[30pt][c]{COCO} & \makebox[30pt][c]{Rel} \\ \midrule
 &  &  & 61.97 & 64.80 & 110.38 & 77.49 \\
\checkmark &  &  & 62.93 & 65.63 & 111.90 & 76.19 \\
 \checkmark & \checkmark &  & 63.06 & 65.89 & 112.37 & 79.42 \\
 \checkmark &  & \checkmark & 63.03 & 67.01 & 112.24 & 80.16 \\
 \checkmark & \checkmark & \checkmark & \textbf{63.48} & \textbf{68.04} & \textbf{112.55} & \textbf{80.69} \\ \bottomrule
\end{tabular}
}
\end{table}

\begin{table}[tbp]\centering
\caption{Ablation study for training strategies. SGE-D denotes the visual relation data used for the training of SGE, while SGE-T denotes the additional training stage for SGE.}  
\label{tab:ablationTraining}
\resizebox{\columnwidth}{!}{
\begin{tabular}{@{}ccc|cccc@{}}
\toprule
SGE-D & SGE & SGE-T & \makebox[15pt][c]{GQA} & \makebox[15pt][c]{MMB} & \makebox[15pt][c]{COCO} & \makebox[15pt][c]{Rel} \\ \midrule
 \checkmark &  &  & 62.32 & 65.72 & 108.45 & 80.23 \\
 \checkmark & \checkmark & & 62.75 & 66.41 & 110.92 & 80.32  \\
 \checkmark & \checkmark & \checkmark & \textbf{63.48} & \textbf{68.04} & \textbf{112.55} & \textbf{80.69} \\ \bottomrule
\end{tabular}
}
\end{table}

\subsection{Overall Performance Assessments}
The comparison results of LLaVA-SG and baseline models are summarized in Table~\ref{tab:result} following the evaluation metrics of \cite{lmms_eval2024}. The reported results of the compared models are collected from the corresponding papers. 
Analyzing the experimental results in Table~\ref{tab:result}, it is evident that our LLaVA-SG model shows significant and consistent improvements over the baseline models. 
The improvements of LLaVA-SG-7B over LLaVA-1.5-7B highlight that our SGE module effectively preserves and expresses visual semantic information with a nearly negligible increase in parameters, achieving significant performance gains. With Vicuna-13B, the LLaVA-SG-13B achieves better results than LLaVA-SG-7B.

The detailed results on MMBench are presented in Table~\ref{tab:result}. MMBench assesses large vision-language models across multiple capability dimensions including LR for Logical Reasoning, AR for Attribute Reasoning, RR for Relation Reasoning, FP-S for Fine-grained Perception (Single Instance), FP-C for Fine-grained Perception (Cross Instance) and CP for Coarse Perception. Analyzing the comparative results in this table, our LLaVA-SG shows significant improvements over LLaVA-1.5 in the capabilities that require entity perception and relationship analysis, specifically in RR, FP-S, FP-C, and CP.

\begin{figure}[tbp]
  \centering
  \includegraphics[width=\columnwidth]{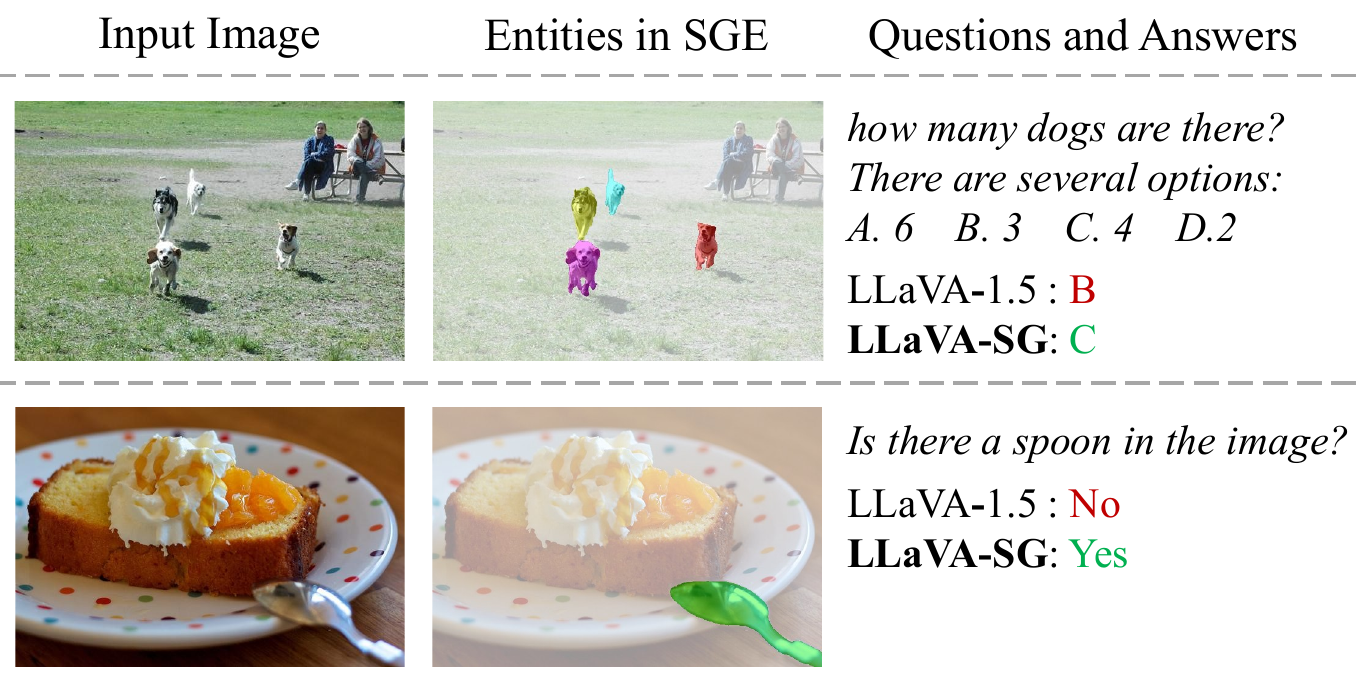}
  \caption{Example outputs of LLaVA-1.5 and our LLaVA-SG model with the first case from MMBench and the second case from POPE. }
  \label{fig:visualization}
\end{figure}

We present example outputs of LLaVA-1.5 and LLaVA-SG in Fig.~\ref{fig:visualization}. 
The middle column shows the masks of entities included in the scene graph of the SGE module. With the visual semantic information expressed by SGE, LLaVA-SG exhibits enhanced multimodal capabilities. For example, in the first case shown in Fig.~\ref{fig:visualization}, without explicitly preserving the visual semantic information in the image, the counting problem becomes difficult for LLaVA. Relying solely on the fragmented visual tokens output by ViT makes it challenging to accurately determine the number of dogs in the image. However, equipped with the SGE module, our LLaVA-SG can leverage tokens that explicitly represent entities in the image, enabling it to provide an accurate count of the dogs.

\subsection{Ablation Study}
We perform ablation experiments to explore the effect of the SGE module and the effect of training strategies. These ablation experiments are based on LLaVA-SG-7B.

\textbf{The Effect of the SGE Module.} We conduct ablation studies on the components within the Scene Graph Expression module. The results are shown in Table~\ref{tab:ablationSGE}. "SG" denotes the construction and use of a basic scene graph without message passing and prompt feature adaptation. "MP" denotes the message passing among nodes in $G$. "Prompt" denotes the attention mechanism with the prompt feature. "Rel" indicates that the test sets from Visual Genome and Open Images are used for evaluating visual relationship classification. As shown in Table~\ref{tab:ablationSGE}, each component contributes to the improvement of the LLaVA baseline.

\textbf{The Effect of Training Strategies.} We performed ablation experiments on the additional visual relationship understanding data used for training the SGE module and the separate SGE training stage. The results are shown in Table~\ref{tab:ablationTraining}. The first row shows the results without using the SGE module, where the visual relationship understanding data used for the second training stage, i.e., the SGE-D in Table~\ref{tab:ablationTraining} is simply appended to the fine-tuning data for training the LLaVA model. The second row shows the results without a separate SGE training stage, where SGE-D is incorporated into the fine-tuning stage and SGE is trained together with the LLM. The third row shows the performance of LLaVA-SG under the full training strategy. The comparison between the first and third rows of Table~\ref{tab:ablationTraining} indicates that the improvement of LLaVA-SG over LLaVA is not primarily due to the additional visual relationship understanding data. Instead, the SGE module plays a crucial role.  The comparison between the second and third rows highlights the necessity of the separate SGE training stage. The separate SGE training stage allows the SGE module to focus more specifically on expressing the visual semantic information within the input images.

\section{Conclusion}
In this paper, we propose a Scene Graph Expression (SGE) module to extract and express visual semantic information structurally for VLM.
With the SGE module, the perception and understanding abilities of VLMs are enhanced.
The LLaVA-SG model, constructed based on the SGE module, shows significant and consistent performance improvements over the baseline methods.

\clearpage


\end{document}